\documentclass[conference]{IEEEtran}
\IEEEoverridecommandlockouts
\usepackage[numbers,square,comma,sort&compress]{natbib}
\usepackage{amsmath,amssymb,amsfonts}
\usepackage{algorithmic}
\usepackage{graphicx}
\usepackage{textcomp}
\usepackage{subfigure}
\usepackage{multirow}
\usepackage{booktabs}
\usepackage{subfigure}
\usepackage{xcolor}
\usepackage{url} 
\usepackage{geometry}
\def\BibTeX{{\rm B\kern-.05em{\sc i\kern-.025em b}\kern-.08em
    T\kern-.1667em\lower.7ex\hbox{E}\kern-.125emX}}

\usepackage{etoolbox}
\makeatletter
\patchcmd{\@makecaption}
  {\scshape}
  {}
  {}
  {}
\makeatletter
\patchcmd{\@makecaption}
  {\\}
  {.\ }
  {}
  {}

\begin{document}

\title{ASY-VRNet: Waterway Panoptic Driving Perception Model based on Asymmetric Fair Fusion of Vision and 4D mmWave Radar}

\author{Runwei Guan$^{1,2,3,4\ \dagger}$, Shanliang Yao$^{1,2,3,4\ \dagger}$, Ka Lok Man$^{3}$, Xiaohui Zhu$^{3}$, Yong Yue$^{3}$, Jeremy Smith$^{1}$, \\ Eng Gee Lim$^{3}$, ~\IEEEmembership{Senior Member,~IEEE}, Yutao Yue$^{5,2,4\ *}$
\thanks{$^{\dagger}$Runwei Guan and Shanliang Yao contribute equally.}
\thanks{

$^{1}$ Department of EEE, University of Liverpool, Liverpool, UK

$^{2}$ Institute of Deep Perception Technology, JITRI, Wuxi, China

$^{3}$ SAT, Xi'an Jiaotong-Liverpool University, Suzhou, China

$^{4}$ XJTLU-JITRI Academy of Industrial Technology, Xi'an Jiaotong-Liverpool University, Suzhou, China

$^{5}$ Thrust of Artificial Intelligence and Thrust of Intelligent Transportation, HKUST (GZ), Guangzhou, China
}
\thanks{$^{*}$ Corresponding author: yutaoyue@hkust-gz.edu.cn}
}

\maketitle

\begin{abstract}
Panoptic Driving Perception (PDP) is critical for the autonomous navigation of Unmanned Surface Vehicles (USVs). A PDP model typically integrates multiple tasks, necessitating the simultaneous and robust execution of various perception tasks to facilitate downstream path planning. The fusion of visual and radar sensors is currently acknowledged as a robust and cost-effective approach. However, most existing research has primarily focused on fusing visual and radar features dedicated to object detection or utilizing a shared feature space for multiple tasks, neglecting the individual representation differences between various tasks. To address this gap, we propose a pair of Asymmetric Fair Fusion (AFF) modules with favorable explainability designed to efficiently interact with independent features from both visual and radar modalities, tailored to the specific requirements of object detection and semantic segmentation tasks. The AFF modules treat image and radar maps as irregular point sets and transform these features into a crossed-shared feature space for multitasking, ensuring equitable treatment of vision and radar point cloud features. Leveraging AFF modules, we propose a novel and efficient PDP model, ASY-VRNet, which processes image and radar features based on irregular super-pixel point sets. Additionally, we propose an effective multitask learning method specifically designed for PDP models. Compared to other lightweight models, ASY-VRNet achieves state-of-the-art performance in object detection, semantic segmentation, and drivable-area segmentation on the WaterScenes benchmark. Our project is publicly available at \textcolor{magenta}{\url{https://github.com/GuanRunwei/ASY-VRNet}}.

% Panoptic Driving Perception (PDP) is crucial for the autonomous navigation of Unmanned Surface Vehicles (USVs). The PDP model is typically a multi-task model, requiring simultaneous and robust completion of multiple perception tasks to facilitate downstream path planning. Currently, the fusion of visual and radar sensors is considered a robust and cost-effective sensor fusion approach. However, most existing research has focused solely on fusing visual and radar features for object detection, neglecting the potential for sharing feature spaces to enhance semantic segmentation performance. In response to this, we propose a pair of asymmetric fusion modules capable of efficiently interacting with independent features from both visual and radar modalities based on the characteristics of object detection and semantic segmentation tasks, fairly treating features of vision and radar point clouds. The asymmetric fair fusion modules transform these features into a shared feature space for multi-tasking. Leveraging these asymmetric fusion modules, we introduce a novel and efficient PDP model, ASY-VRNet, which fairly treats image and radar features as irregular point sets. Furthermore, we propose an effective multi-task learning method tailored for PDP models. Based on the aforementioned, compared with other lightweight models, our ASY-VRNet achieves state-of-the-art performance on object detection, semantic segmentation and drivable-area segmentation on the WaterScenes benchmark. 
\end{abstract}

\begin{IEEEkeywords}
waterway panoptic driving perception, vision-radar fusion, asymmetric fusion, multi-task learning
\end{IEEEkeywords}

\section{Introduction}
With the rapid and exhilarating development of artificial intelligence and sophisticated perception sensors, USVs demonstrate immensely promising value in channel monitoring, water-quality assessment, water-surface rescue operations, water-surface transportation, and geological prospecting \cite{yao2023radar,guan2024mask}. As one of the most crucial and foundational modules, perception is vital for the autonomous and efficient navigation of USVs. Currently, Panoptic Driving Perception (PDP) is regarded as an extraordinarily effective paradigm for comprehensive environmental perception, typically relying on advanced vision sensors. PDP operates based on multi-task robust perception for different driving areas, aiming at the simultaneous instance-level perception of objects and pixel-level recognition of drivable areas \cite{wu2022yolop,han2022yolopv2,vu2022hybridnets}, which facilitates the comprehensive understanding of the environment. In contrast to integrating multiple single-task models, PDP models can significantly reduce memory usage and dramatically enhance inference speed. Moreover, through meticulously designed and well-coordinated multi-task training patterns, they effectively improve the performance of individual tasks, demonstrating both impressive efficiency and remarkable accuracy.

\begin{figure}
    \centering
    \includegraphics[width=0.99\linewidth]{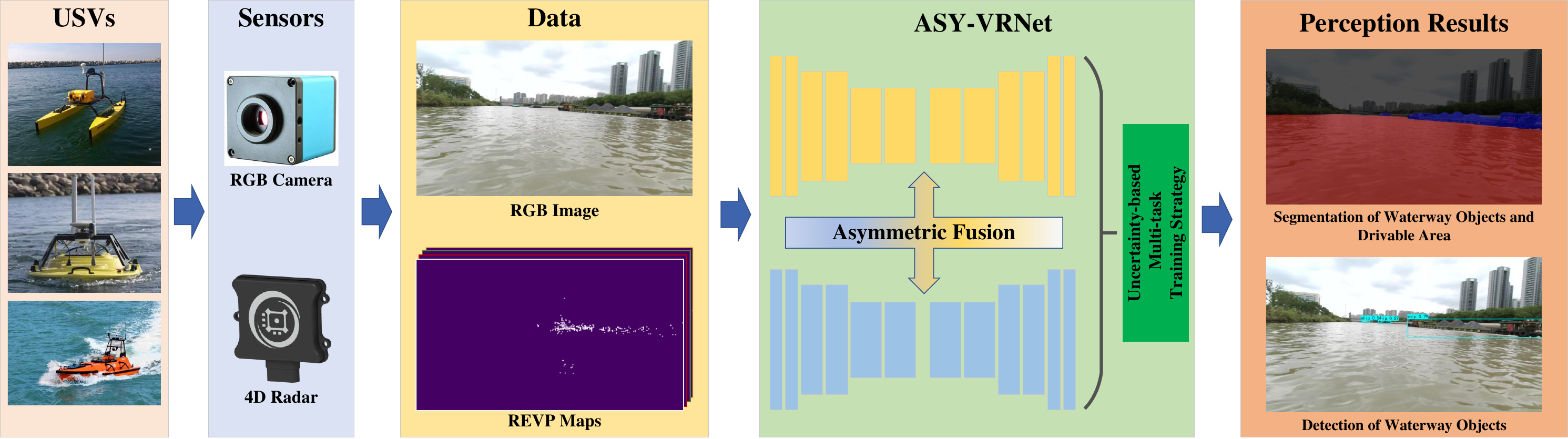}
    \vspace{-5mm}
    \caption{The overview of our proposed methods. It contains five parts, USVs, sensors (monocular camera and 4D radar), data perceived by sensors, ASY-VRNet, multi-task training strategy and perception results.}
    \label{fig:overview}
\end{figure}

\begin{figure}
    \centering
    \includegraphics[width=0.99\linewidth]{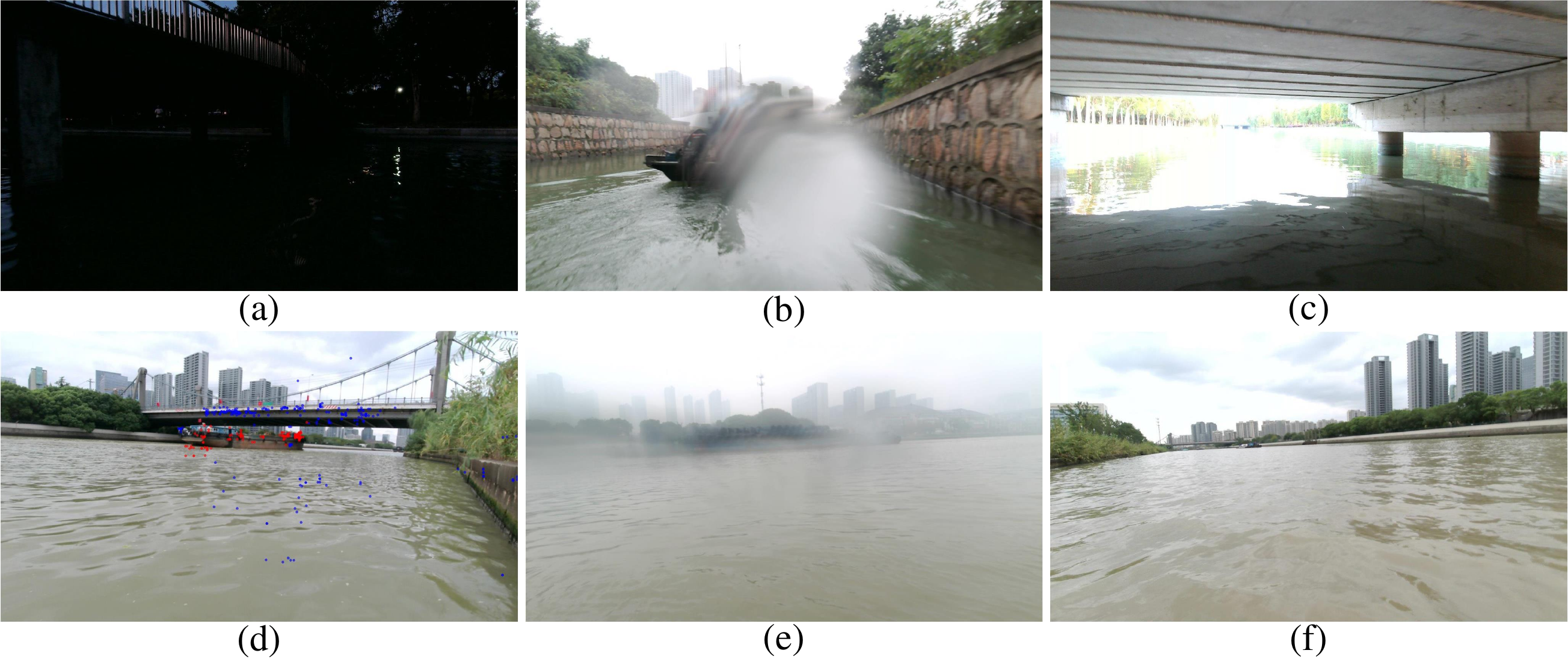}
    \vspace{-5mm}
    \caption{Several challenging scenes in waterway perception: (a) dark environment, (b) camera malfunction, (c) strong light, (d) radar clutter, (e) adverse weather and (f) small objects.}
    \label{fig:several_scenes}
\end{figure}

However, as illustrated in Fig. \ref{fig:several_scenes}, purely visual solutions often prove unreliable in various aquatic environments, such as (a) low-light conditions, (b) water droplets on the lens, (c) strong reflections on the water surface, (e) dense water fog, and (f) small objects. Currently, the fusion of visual information with 4D millimeter-wave radar (4D radar) is regarded as a promising, reliable, and cost-effective approach. As an all-weather perception sensor, 4D radar remains unaffected by adverse weather conditions and provides denser point cloud information compared to 3D radar, despite being susceptible to multi-path clutter (Fig. \ref{fig:several_scenes} (d)). Numerous studies have explored feature-level fusion of radar and vision for environmental perception, primarily aiming to enhance object detection performance \cite{guan2023achelous, guan2024mask, guan2023achelous++, cheng2021robust}. Nevertheless, enabling region-level object detection and pixel-level semantic segmentation tasks through the effective utilization of vision and radar deep features to mutually enhance their performance remains a challenge. Specifically, radar point clouds are typically sparse and irregular, and visual objects often exhibit similar irregularities, suggesting that conventional convolution-based modules, which rely on 2D rectangular structures, may not effectively model both modalities. Additionally, region-level detection and pixel-level segmentation have different feature representation and optimization requirements, where the shared fusion modules \cite{guan2023achelous, guan2023achelous++, guan2024mask} usually cannot help these two types of tasks achieve their respective optimal performances. Therefore, we propose a cephalocaudal feature structure that can impartially and consistently treat visual and radar point cloud features as sets of irregular pixel points during feature extraction, alignment, and fusion. Additionally, we design two specialized vision-radar fusion modules aimed at maximizing the performance of both detection and segmentation tasks.

\begin{figure*}
    \centering
    \includegraphics[width=0.99\linewidth]{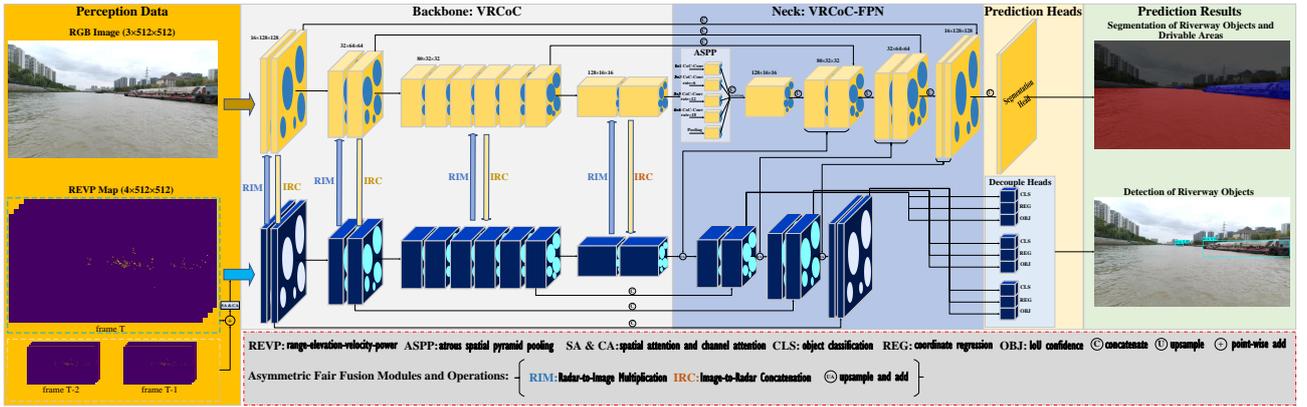}
    \vspace{-3mm}
    \caption{The architecture of our proposed ASY-VRNet. It contains five parts, perception data, VRCoC, VRCoC-FPN, prediction heads and Asymmetric Fair Fusion modules (AFF), including RIM and IRC. Each stage of VRCoC has 2, 2, 6, 2 stacking blocks. VRCoC, AFF and VRCoC-FPN (Feature Pyramid Network) are three dedicated designed components in this paper.}
    \label{fig:neural_network}
\end{figure*}

Building on above, we concentrate on robust and high-performance panoptic driving perception for waterway autonomous driving and our contributions are as follows:

\begin{enumerate}
    \item We design and propose a robust model for waterway panoptic driving perception named ASY-VRNet. ASY-VRNet utilizes a full Contextual-Clustering (CoC) architecture, including both the backbone and neck, treating image and radar objects equitably as irregular point sets.
    \item Drawing from prior and theoretical principles, we develop a pair of effective fusion methods for vision-radar integration, termed Asymmetric Fair Fusion (AFF). AFF takes into account the distinct characteristics of different perception tasks and designs the fusion processes based on the unique features of images and radar maps, ensuring an unbiased approach to both modalities. AFF enhances explainability and can be employed as a plug-and-play module to improve the performance of any vision-radar fusion network.
    \item Inspired by homoscedastic uncertainty \cite{kendall2018multi}, we devise a multi-task training strategy that leverages the inherent uncertainty of perception tasks during the training of PDP models.
\end{enumerate}

The remainder of this paper is organized as follows: Section \ref{sec:related_works} reviews related work; Section \ref{sec:asy_model} details our proposed ASY-VRNet model; Section \ref{sec:experiments} presents the experimental setup and results; and Section \ref{sec:conclusions} concludes our study.

\section{Related Works}
\label{sec:related_works}

\subsection{Panoptic Driving Perception in Autonomous Driving}
Panoptic driving perception (PDP) is crucial in autonomous driving systems for UGVs, USVs, and UAVs. PDP models are primarily responsible for obstacle detection and drivable area segmentation to facilitate downstream path planning. Current PDP models can be categorized into vision-based and fusion-based models. Vision-based models, such as YOLOP \cite{wu2022yolop} and YOLOPv2 \cite{han2022yolopv2}, both based on YOLO architecture, are capable of detecting traffic participants, identifying lanes, and segmenting drivable areas simultaneously. HybridNets \cite{vu2022hybridnets}, utilizing EfficientNet \cite{tan2019efficientnet} as the backbone, combines this with an effective multi-task training strategy. In waterway perception, sensor-based models primarily involve the fusion of vision and radar. Mask-VRDet \cite{guan2024mask} employs a dual graph fusion (DGF) method to integrate image and radar features. Achelous \cite{guan2023achelous} and Achelous++ \cite{guan2023achelous} can perform five tasks concurrently, adopting both feature-level and proposal-level fusion strategies.

\subsection{Multi-Task Learning Strategies in Computer Vision}
Multi-task learning (MTL) is a fundamental and challenging problem in deep learning. Balancing the loss of various tasks during training based on their loss values and characteristics presents an intriguing proposition. For related tasks, designing appropriate MTL strategies can effectively enhance the representation of shared features, thereby improving performance across different tasks. GradNorm \cite{chen2018gradnorm} is a classical MTL method that scales the loss of different tasks to a similar magnitude. Dynamic Weight Averaging (DWA) \cite{liu2019end} balances the learning paces of various tasks. Sener \textit{et al.} \cite{sener2018multi} approach MTL as a multiple objective optimization (MOO) problem, finding the Pareto optimal solution among various tasks. Kendall \textit{et al.} \cite{kendall2018multi} propose an uncertainty-based MTL method, estimating the weight of various tasks through Gaussian likelihood estimation.

\subsection{Simple Linear Iterative Clustering Algorithm and Its Extension }
The Simple Linear Iterative Clustering (SLIC) algorithm \cite{achanta2012slic} generates superpixels with perceptual significance through \textit{k}-means clustering. Superpixels provide a flexible alternative to the conventional pixel grid structure. Building on SLIC, Ma \textit{et al.} \cite{ma2023image} propose Contextual Clustering (CoC), an image feature extractor that treats an image as a set of points. CoC exhibits promising potential and generalizability for multi-modal learning involving images and point clouds.

\section{ASY-VRNet}
\label{sec:asy_model}
This section presents a detailed exposition of the design of ASY-VRNet, organized into six parts corresponding to the major modules of the model.

\subsection{Alignment and Pre-processing of Perception Data}
Given that all tasks are based on the camera plane, we initially project the 3D radar point cloud onto the camera plane through a coordinate system transformation.

After projecting the radar point clouds onto the camera plane, each radar point encapsulates features such as the object's range, elevation, velocity, and reflected power. Leveraging these characteristics, we design a radar data representation known as REVP maps, which is a 4-channel image-like feature map.

\subsection{Vision-Radar-based Contextual Clustering (VRCoC)}

\begin{figure}
    \centering
    \includegraphics[width=0.80\linewidth]{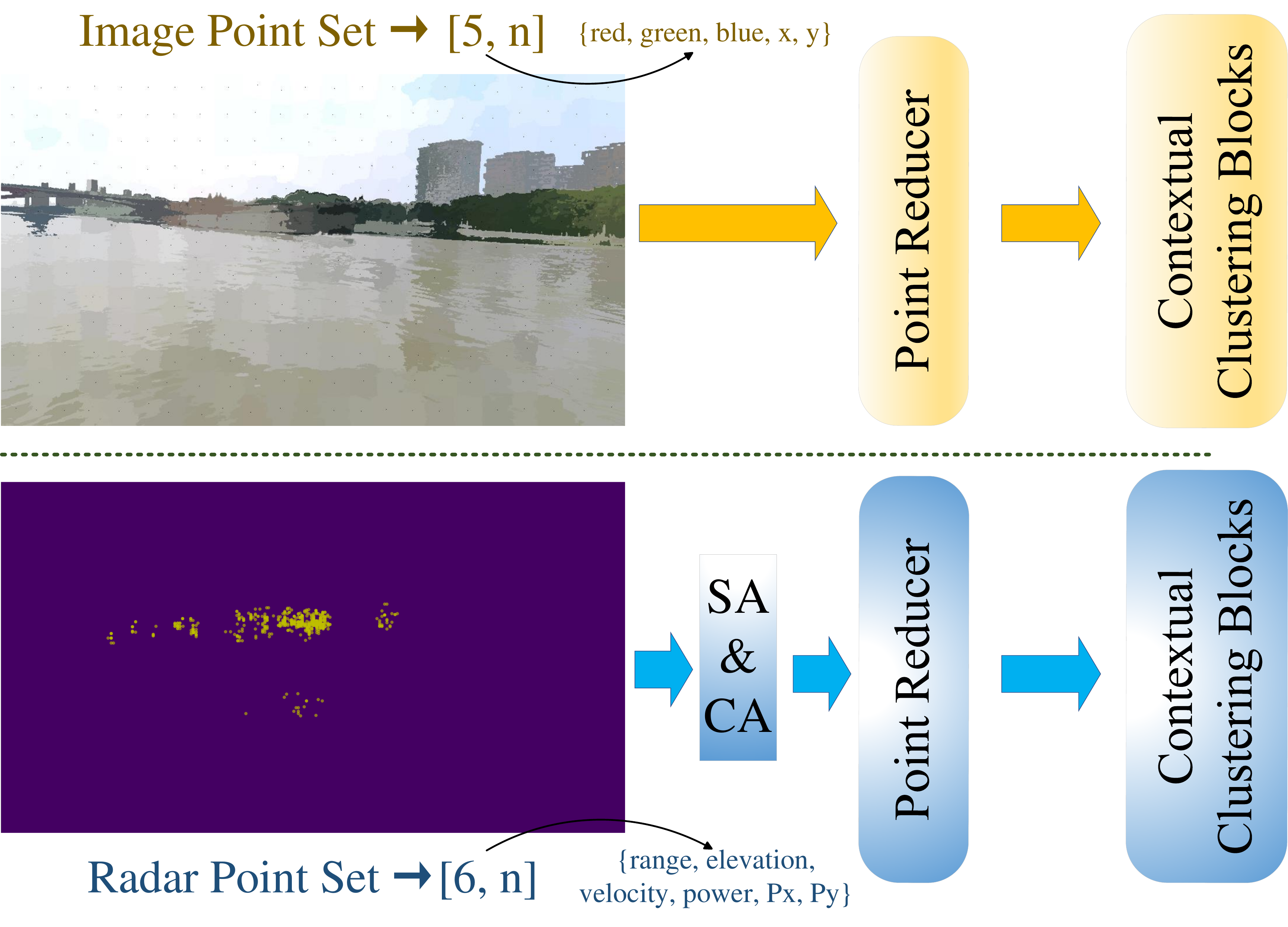}
    \vspace{-3mm}
    \caption{The first stage of VRCoC, including image-like point sets (image and radar), point reducer and contextual clustering blocks.}
    \label{fig:coc_vis}
\end{figure}

\textbf{V}ision-\textbf{R}adar-based \textbf{Co}ntextual \textbf{C}lustering (VRCoC) constitutes the backbone of ASY-VRNet, facilitating hierarchical feature extraction across four stages. As illustrated in Fig. \ref{fig:neural_network}, VRCoC is a dual-branch backbone. Each branch consists of individual blocks formed by the combination and stacking of two fundamental modules: Point Reducer and Contextual Clustering. These branches are specifically designed to extract features from images and radar REVP maps, respectively.

\textbf{Feature Preparation.} Given an image \(I \in \mathbb{R}^{3 \times w \times h}\) and a REVP map \(R \in \mathbb{R}^{4 \times w \times h}\), we assign the coordinate of each pixel in the RGB image \(I_{i,j}\) as \(\left[ \frac{i}{w} - 0.5, \frac{j}{h} - 0.5 \right]\). Similarly, the coordinate of each element in the REVP map \(R_{i,j}\) is assigned as \(\left[ \frac{i}{w} - 0.5, \frac{j}{h} - 0.5 \right]\). Consequently, the RGB image is transformed into a set of points \(IP \in \mathbb{R}^{5 \times n}\) while the REVP map is transformed into a set of points \(RP \in \mathbb{R}^{6 \times n}\), where \(n = w \times h\) is the point count per channel in both the RGB image and the REVP map. Each point in the RGB image encompasses color features (3 channels) and positional features (2 channels). For the REVP map, each point includes radar-captured object features (4 channels) and positional features (2 channels). Due to radar interference from multipath effects and surface clutter, numerous non-object clutter points may appear. To mitigate the impact of clutter features on model optimization, we incorporate dual attention mechanisms for both spatial and channel dimensions before feeding the REVP map into the Point Reducer module (Fig. \ref{fig:coc_vis}). This allows the model to adaptively adjust its focus on different positions and feature channels of the REVP map. Our channel attention is based on the Efficient Channel Attention (ECA) module \cite{wang2020eca}, while the spatial attention utilizes Deformable Convolution \cite{zhu2019deformable}, effectively modeling the irregular features of point clouds.

\textbf{Point Reducer.} Based on the sets of points \(IP \in \mathbb{R}^{5 \times n}\) and \(RP \in \mathbb{R}^{6 \times n}\) obtained, we proceed with feature extraction. As depicted in Fig. \ref{fig:coc_vis}, the first step involves the Point Reducer, which reduces the number of points. In this step, following the Contextual Clustering (CoC) methodology \cite{ma2023image}, we evenly select anchors in the feature space and concatenate the nearest \(k\) centers of points. Subsequently, a linear feed-forward module is employed to transform the feature map dimensions to \(d\).

\textbf{Contextual Clustering.} 
Based on the feature points of the image \(IP \in \mathbb{R}^{d \times n}\) and the REVP map \(RP \in \mathbb{R}^{d \times n}\) at the same stage, we group feature points into several clusters based on the cosine similarity between the features of the points and the clustering centers. The clustering centers are evenly selected using the SLIC algorithm \cite{achanta2012slic}. After assigning points to their respective centers, feature aggregation is applied upon the similarities between the clustering points and the clustering center. Assuming there are \(m\) clustering points in a cluster, the similarity matrix between the clustering points and the clustering center is denoted as \(s \in \mathbb{R}^m\). We then map these feature points into an aggregation space, so \(IP \in \mathbb{R}^{d \times m}\) and \(RP \in \mathbb{R}^{d \times m}\) are transformed to \(IP \in \mathbb{R}^{\hat{d} \times m}\) and \(RP \in \mathbb{R}^{\hat{d} \times m}\), respectively, where \(\hat{d}\) is the dimension of the feature points in the aggregation space. Within each cluster in the aggregation space, a clustering center \(v_c\) is similarly proposed based on the SLIC algorithm. Therefore, the aggregated feature of points \(f \in \mathbb{R}^{\hat{d} \times m}\) is presented in Equation \ref{eq:aggregate-feature}.

\begin{equation}
\begin{aligned}
 &   f = \frac{1}{C}\Big(v_c + \sum\limits_{i=1}^{m}\sigma(\alpha s_i + \beta) * v_i\Big), \\
 &  s.t., C=1+\sum\limits_{i=1}^{m} \sigma (\alpha s_i + \beta), 
 \label{eq:aggregate-feature}
\end{aligned}
\end{equation}
where $\alpha$ and $\beta$ are learnable parameters representing the scale and shift ratio of the similarity. $\sigma$ is the sigmoid function to scale the similarity to $(0,1)$. $v_i$ denotes the $i$th points in the aggregation space. $C$ is a normalization factor.

Then the aggregated feature $f$ is dispatched to each feature point in the cluster according to the similarity. For each feature point $p_i$, the dispatch step for updating is presented in Equation \ref{eq:dispatch-feature},

\begin{equation}
 \hat{p_i} = p_i + FF\Big(\sigma(\alpha s_i + \beta)* f\Big), \quad \hat{p_i} \in \mathbb{R}^{d \times n},
  \label{eq:dispatch-feature}
\end{equation}
where $p_i$ represents the $i^{th}$ feature point and $\hat{p_i}$ represents the updated $i^{th}$ feature point. $\sigma$ is the sigmoid function. $FF$ is the feed-forward module based on fully-connected layers, which transforms the dimension $\hat{d}$ back to $d$.

Based on the above, we obtain the feature maps of the image $\hat{IP}$ and radar $\hat{RP}$ updated by the point reducer and contextual clustering in CoC. Inspired by multi-head self-attention, we divide the channel of $\hat{IP}$ and $\hat{RP}$ into $h$ parts, and each part denotes one head. Each head $head_i$ is weighted individually and concatenated to other heads. After that, we concatenate all heads along the dimension of channels. Multi-head operation can make the network adaptively attach importance to features. The process of the multi-head operation is shown in Equation \ref{eq:multi-head-coc}.

\begin{equation}
\footnotesize
\begin{aligned}
\label{eq:multi-head-coc}
 \hat{IP}^{'} & = [head^{IP}_1 W^{IP}_1, head^{IP}_2 W^{IP}_2,  \dots , head^{IP}_h W^{IP}_h], \\
 \hat{RP}^{'} & = [head^{RP}_1 W^{RP}_1, head^{RP}_2 W^{RP}_2,  \dots , head^{RP}_h W^{RP}_h].
\end{aligned}
\end{equation}

\subsection{Asymmetric Fair Fusion Modules}
The Asymmetric Fair Fusion (AFF) modules are a set of bidirectional fusion mechanisms designed for the mutual integration and enhancement of visual and radar features. They are structured asymmetrically and consist of two primary components: Image-Radar Concatenation (IRC) and Radar-Image Multiplication (RIM). 

\begin{figure}[ht]
    \centering
    \includegraphics[width=0.97\linewidth]{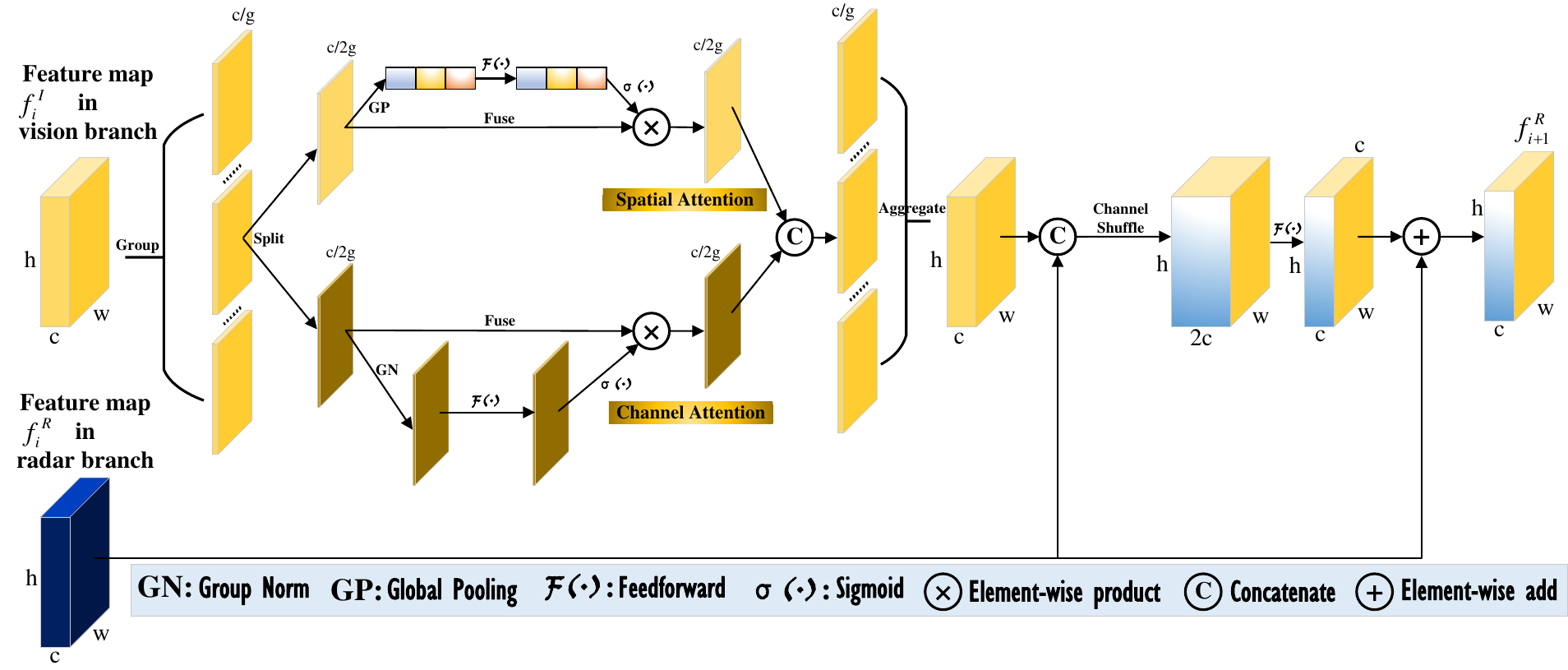}
    \vspace{-4mm}
    \caption{The structure of Image-Radar Concatenation (IRC).}
    \label{fig:irc}
\end{figure}

\textbf{Image-Radar Concatenation (IRC)} is designed to enhance object detection representation. It is widely recognized that image-based object detection results are not directly influenced by pixel brightness or color. For instance, a luminous area does not necessarily indicate the location of an object. Neural network models typically learn from a vast number of images to identify various feature combinations for object localization. However, 4D radar significantly improves this process. 4D radar can capture denser point clouds of objects than ordinary radar, regardless of whether the object is moving or stationary. This capability allows the point cloud of 4D radar to help the neural network model anchor the general area of the object early in training, thereby accelerating the convergence of object detection. In adverse weather and low-light environments, the 4D radar point cloud can compensate for the lack of visual features, reducing the likelihood of missed detections.

Based on the aforementioned principles, we propose the Image-to-Radar Concatenation (IRC) module. As depicted in Fig. \ref{fig:irc}, let's consider a feature map \( f^I_i \in \mathbb{R}^{c \times h \times w} \) in the vision branch and a feature map \( f^R_i \in \mathbb{R}^{c \times h \times w} \) in the radar branch. \( f^I_i \) is initially divided into \( g \) segments, where each segment is denoted as \( f^I_{i_j} \in \mathbb{R}^{\frac{c}{g} \times h \times w} \). These segments undergo further processing: one branch focuses on spatial attention while the other on channel attention, following the principles of shuffle attention \cite{zhang2021sa}.

For channel attention, given the input feature map \( f^I_{i_{j-c}} \in \mathbb{R}^{\frac{c}{2g} \times h \times w} \), as expressed in Equation \ref{eq:irc-ca}, \( f^I_{i_{j-c}} \) undergoes global average pooling to capture its global representation. Subsequently, non-linear and sigmoid functions are applied to assess the importance of each channel. Finally, the channel importance weight is applied to \( f^I_{i_{j-c}} \), resulting in the feature map with channel attention \( \hat{f}^I_{i_{j-c}} \in \mathbb{R}^{\frac{c}{2g} \times h \times w} \).

\begin{equation}
\footnotesize
\begin{aligned}\label{eq:irc-ca}
 & f^I_{i_{j-c-1}} = \sigma \Big(W_c \cdot GP(f^I_{i_{j-c}}) + b_c \Big), \quad f^I_{i_{j-c-1}} \in \mathbb{R}^{\frac{c}{2g} \times 1 \times 1}, \\
 & \hat{f}^I_{i_{j-c}} = f^I_{i_{j-c-1}} * f^I_{i_{j-c}}, \quad \hat{f}^I_{i_{j-c}} \in \mathbb{R}^{\frac{c}{2g} \times h \times w},
\end{aligned}
\end{equation}
where $GP$ denotes global average pooling. $W_c$ is the learnable weight in the non-linear feed-forward module while $b_c$ is the learnable bias. $\sigma$ is the sigmoid function. $*$ denotes element-wise multiplication.

For the spatial attention, given the input feature map $f^I_{i_{j-s}} \in \mathbb{R}^{\frac{c}{2g} \times h \times w}$, as Equation \ref{eq:irc-sa} presents, $f^I_{i_{j-s}}$ is first normalized by group, then processed by a non-linear feed-forward module and a sigmoid function to measure the spatial importance. Finally, the spatial importance is multiplied with the input feature map $f^I_{i_{j-s}}$ to obtain the feature map with spatial attention $\hat{f}^I_{i_{j-s}} \in \mathbb{R}^{\frac{c}{2g} \times h \times w}$.

\begin{equation}
\footnotesize
\begin{aligned}\label{eq:irc-sa}
 & f^I_{i_{j-s-1}} = \sigma \Big(W_s \cdot GN(f^I_{i_{j-s}}) + b_s \Big), \quad f^I_{i_{j-s-1}} \in \mathbb{R}^{\frac{c}{2g} \times h \times w}, \\
 & \hat{f}^I_{i_{j-s}} = f^I_{i_{j-s-1}} * f^I_{i_{j-s}}, \quad \hat{f}^I_{i_{j-s}} \in \mathbb{R}^{\frac{c}{2g} \times h \times w}, 
\end{aligned}
\end{equation}
where $GN$ denotes group-norm. $W_s$ is the learnable weight in the non-linear feed-forward module while $b_s$ is the learnable bias. $\sigma$ is the sigmoid function while $*$ denotes element-wise multiplication.

After that, we concatenate $\hat{f}^I_{i_{j-s}}$ and $\hat{f}^I_{i_{j-c}}$ and get the combination of the feature map $\hat{f}^I_{i_{j-sc}}$ with both channel and spatial attention. Aggregation (concatenation) of $g$ $\hat{f}^I_{i_{j-sc}}$ is implemented to get the initial feature map with channel and spatial attention $f^I_{i_{sc}} \in \mathbb{R}^{c \times h \times w}$. We concatenate $f^I_{i_{sc}}$ and $f^R_i \in \mathbb{R}^{c \times h \times w}$ along the channel dimension (Equation \ref{eq:irc-concat-ir}).

\begin{equation}\label{eq:irc-concat-ir}
    f^{IR}_i = [f^I_{i_{sc}}, f^R_i], \quad f^{IR}_i \in \mathbb{R}^{2c \times h \times w},
\end{equation}
where $[\cdot]$ is the concatenation operation.

After that, the channel shuffle is exerted to enhance the interaction among features, followed by a feed-forward module, reducing the channel dimension. Finally, a long residual path is added and we get the feature map $f^{R}_{i+1}$ updated by \textit{IRC} in the radar branch. The whole process is shown in Equation \ref{eq:irc-final-stage}.

\begin{equation}
\begin{aligned}\label{eq:irc-final-stage}
 f^{IR}_i & = S(f^{IR}_i), \quad f^{IR}_i \in \mathbb{R}^{2c \times h \times w}, \\
 \hat{f}^{IR}_i & = W_f \cdot f^{IR}_i + b_f, \quad \hat{f}^{IR}_i \in \mathbb{R}^{c \times h \times w}, \\
 f^{R}_{i+1} & = \hat{f}^{IR}_i + f^R_i, \quad f^{R}_{i+1} \in \mathbb{R}^{c \times h \times w},
\end{aligned}
\end{equation}
where $S(\cdot)$ denotes the channel shuffle. $W_f$ is the learnable weight in the non-linear feed-forward module while $b_f$ is the learnable bias.

\textbf{Radar-Image Multiplication (RIM)} is to enhance the representation for image segmentation as radar point clouds can be seen as sparse masks of objects. RIM is based on the formula of brightness and contrast adjustment (Equation \ref{eq:bright-adjust}).

\begin{equation}
    g(i, j) = \alpha f(i, j) + \beta,
    \label{eq:bright-adjust}
\end{equation}
where $f(i, j)$ is the pixel in the original image while $g(i, j)$ is the pixel after adjustment. $\alpha$ is the gain to adjust the image contrast while $\beta$ is the bias to control the image brightness. Based on the above, we intend to use features in the radar branch to focus on and enhance the features in the vision branch at same positions.

\begin{figure}[h]
    \centering
    \includegraphics[width=0.86\linewidth]{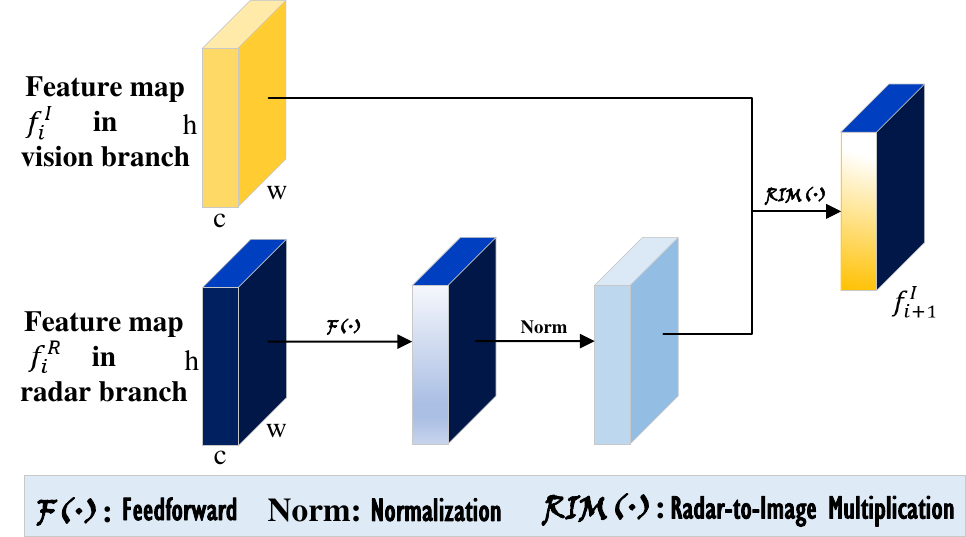}
    \vspace{-3mm}
    \caption{The architecture of Radar-to-Image Multiplication (RIM).}
    \label{fig:rim}
\end{figure}

% Fig. \ref{fig:rim} presents the architecture of RIM. Assuming a feature map $f^I_i \in \mathbb{R}^{c \times h \times w}$ in the vision branch and a feature map $f^R_i \in \mathbb{R}^{c \times h \times w}$ in the radar branch. As Equation \ref{eq:rim} presents, $f^R_i$ is first through a feed-forward module and then normalized to the feature map $\hat{f}^R_i$. Based on Equation \ref{eq:bright-adjust}, here $\alpha = 1 + \hat{f}^R_i$ and $\beta = \gamma * \hat{f}^R_i$ are used to enhance the feature of the corresponding positions in the vision branch. Finally, we get the image feature map $f^I_{i+1}$, which is an image feature map containing the attention from the radar feature map.
Fig. \ref{fig:rim} presents the architecture of the Radar-Image Multiplication (RIM) module. Assuming a feature map $f^I_i \in \mathbb{R}^{c \times h \times w}$ in the vision branch and a feature map $f^R_i \in \mathbb{R}^{c \times h \times w}$ in the radar branch, as shown in Equation \ref{eq:rim}, $f^R_i$ first undergoes a feed-forward module and normalization to produce the feature map $\hat{f}^R_i$. Following Equation \ref{eq:bright-adjust}, the parameters $\alpha = 1 + \hat{f}^R_i$ and $\beta = \gamma * \hat{f}^R_i$ are utilized to enhance the corresponding positions in the vision branch. This process results in the image feature map $f^I_{i+1}$, which incorporates cross attention from the radar feature map.

\begin{equation}
\begin{aligned}
 \hat{f}^R_i & = Norm(W_r \cdot f^R_i + b_r), \quad \hat{f}^R_i \in \mathbb{R}^{c \times h \times w}, \\
 f^I_{i+1} & = (1 +\hat{f}^R_i) \cdot f^I_i + \gamma * \hat{f}^R_i, \quad f^I_{i+1} \in \mathbb{R}^{c \times h \times w},
 \label{eq:rim}
\end{aligned}
\end{equation}
where $Norm$ is the normalization operation. $W_r$ is the learnable weight and $b_r$ is the learnable bias in the feed-forward module. $\gamma$ is a learnable coefficient.

\subsection{Dual Feature Pyramid Networks}

To maintain consistency in the FPN structure with the backbone, we continue to employ Contextual Clustering (CoC) as the fundamental unit across all stages of FPN, termed VRCoC-FPN. Illustrated in Fig. \ref{fig:neural_network}, VRCoC-FPN retains the dual-branch architecture akin to VRCoC. In the vision branch, each stage incorporates an Atrous Spatial Pyramid Pooling (ASPP) module \cite{chen2018encoder} to enhance the receptive field across multiple scales. Skip connections are employed within VRCoC-FPN to facilitate multi-scale feature fusion. Meanwhile, in the radar branch, each stage integrates feature maps from the corresponding stage in the vision branch to enhance resolution for object detection.

\subsection{Predictions Heads}
The model comprises two distinct prediction heads: one for semantic segmentation and another for object detection. The segmentation head consists of $C_{seg}+1$ channels, where $C_{seg}$ denotes the number of semantic segmentation categories, and $1$ represents the background. For object detection, we utilize decoupled heads, inspired by YOLOX \cite{ge2021yolox}, to independently predict the bounding box coordinates, object category, and confidence score. Additionally, ASY-VRNet is anchor-free and employs SimOTA \cite{ge2021yolox} for dynamic positive sample matching.

\subsection{Multi-task Optimization Strategy}

Given the significant disparity in loss magnitudes between object detection and semantic segmentation, we adopt a multi-task loss approach inspired by Kendall \textit{et al.} \cite{kendall2018multi}, which is based on homoscedastic uncertainty. Homoscedastic uncertainty, a subset of aleatoric uncertainty, pertains to inherent data randomness and unexplainable information. ASY-VRNet addresses two primary tasks: object detection and semantic segmentation. Object detection includes the loss of bounding box coordinates, confidence and object category, which are one regression and two classification tasks. Semantic segmentation is a pixel-level classification task. Therefore, we can consider these two tasks as a combination of regression and classification sub-tasks. We define the loss of our ASY-VRNet model as $L(W, \sigma_1, \sigma_2, \sigma_3, \sigma_4)$, which can be written as,

\begin{equation}
\begin{aligned}
    & L(W, \sigma_1, \sigma_2, \sigma_3, \sigma_4) \\
    & \hspace{-5mm} = \sum \limits_{i=1}^3 \frac{1}{\sigma_i^2}L_i(W) + \frac{1}{\sigma_4^2}L_4(W) + \sum \limits_{k=1}^4 log\sigma_k,
    \label{eq:hu-mtl}
\end{aligned}
\end{equation}
where $\sigma_1$, $\sigma_2$, $\sigma_3$, $\sigma_4$ respectively represent the uncertainty of the data for 4 sub-tasks in panoptic perception: object classification, object confidence score, pixel classification and bound box regression. $log \sigma _k$ is the regularization term. If $\sigma_k$ became larger, the weight of $L_k(W)$ would be smaller. In practical training, the uncertainty of the input data distribution always exists, which means $\sigma_k$ is positive and will not be zero.

\section{Experiments}
\label{sec:experiments}

\subsection{Experimental Settings}
We train and evaluate ASY-VRNet on the WaterScenes dataset \cite{yao2023waterscenes}, including 54,120 frames and seven categories in various waterway scenarios. We train all models in experiments for 100 epochs with a batch size of 16. We adopt Stochastic Gradient Descent with Momentum (SGDM) as the optimizer. The weight decay is 5e-4 while the momentum is 0.937. We adopt a cosine learning rate scheduler with an initial learning rate of 1e-2. In addition, we adopt the weather augmentation modules in Albumentations \cite{info11020125} to simulate different adverse weather. Both images and REVP maps are resized as $320 \times 320$ (px) during the training.
Furthermore, we adopt Exponential Moving Average (EMA) and Mixed Precision (MP). For the test, we choose mAP, AP and AR as metrics to evaluate the detection performances, while mIoU is the semantic segmentation metric. All training and test works are implemented on one TITAN RTX GPU.

\begin{table}
\setlength\tabcolsep{0.8pt} 
\caption{Comparison of ASY-VRNet with Other Models on Object Detection}
\vspace{-3mm}
\centering
\label{tab:detection_compare_map}
\begin{tabular}{cccccc}  
\toprule   
  \textbf{Models} & \textbf{Modalities} & \textbf{Params (M)} & \textbf{FLOPs (G)} & \textbf{mAP}$_\text{50-95}$ & \textbf{AR}$_\text{50-95}$ \\
\midrule   
\multicolumn{6}{c}{\texttt{Single-Task Models}} \\
\midrule
  YOLOv4-T \cite{ge2021yolox} & $\mathbb{V}$ & 5.89 & 4.04 & 13.1 & 20.2 \\
  YOLOv7-T \cite{wang2023yolov7} & $\mathbb{V}$ & 6.03 & 33.3 & 37.3 & 43.7 \\
  YOLOX-T \cite{ge2021yolox} & $\mathbb{V}$ & 5.04 & 3.79 & 39.4 & 43.0 \\
  YOLOv8-N \cite{yolov8_ultralytics}  & $\mathbb{V}$ & 3.01 & 2.05 & 41.9 & 44.0 \\ 
  CRFNet \cite{nobis2019deep} & $\mathbb{V}$+$\mathbb{R}$ & 23.54 & - & 41.8 & 44.5 \\
\midrule
\multicolumn{6}{c}{\texttt{Multi-Task Models}} \\
\midrule
  YOLOP \cite{wu2022yolop} & $\mathbb{V}$ & 7.90 & 18.60 & 37.9 & 43.5 \\
  HybridNets \cite{vu2022hybridnets} & $\mathbb{V}$ & 12.83 & 15.60 & 39.1 & 44.2 \\
  Achelous \cite{guan2023achelous} & $\mathbb{V}$+$\mathbb{R}$ & 3.49 & 3.04 & 41.5 & 45.6 \\
\midrule
  \textbf{ASY-VRNet} & \textbf{$\mathbb{V}$+$\mathbb{R}$} & \textbf{4.12} & \textbf{3.26} & \textbf{42.8} & \textbf{46.3} \\
\bottomrule 
\end{tabular}
\end{table}

\begin{table}
\setlength\tabcolsep{1.0pt} 
\caption{Comparison of ASY-VRNet with Other Models on Semantic Segmentation (Object and Drivable-area)}
\vspace{-3mm}
\centering
\label{tab:detection_compare_miou}
\begin{tabular}{cccccc} 
\toprule   
  \textbf{Models} & \textbf{Modalities} & \textbf{Params (M)} & \textbf{FLOPs (G)} & \textbf{mIoU$_{\text{o}}^1$} & \textbf{mIoU$_{\text{d}}^2$} \\
\midrule   
\multicolumn{6}{c}{\texttt{Single-Task Models}} \\
\midrule
  Segformer-B0 \cite{xie2021segformer} & $\mathbb{V}$ & 3.71 & 5.29 & 73.5 & 99.4 \\
  DeepLabV3+ \cite{Chen_Zhu_Papandreou_Schroff_Adam_2018} & $\mathbb{V}$ & 5.81 & 20.60 & 71.6 & 99.2 \\
  PSPNet \cite{zhao2017pyramid} & $\mathbb{V}$ & 2.38 & 2.30 & 69.4 & 99.0 \\
\midrule
\multicolumn{6}{c}{\texttt{Multi-Task Models}} \\
\midrule
  YOLOP \cite{wu2022yolop} & $\mathbb{V}$ & 7.90 & 18.60 & - & 99.0 \\
  HybridNets \cite{vu2022hybridnets} & $\mathbb{V}$ & 12.83 & 15.60 & - & 98.8 \\
  Achelous \cite{guan2023achelous} & $\mathbb{V}$+$\mathbb{R}$ & 3.49 & 3.04 & 70.6 & 99.5 \\
\midrule
  \textbf{ASY-VRNet} & \textbf{$\mathbb{V}$+$\mathbb{R}$} & \textbf{4.12} & \textbf{3.26} & \textbf{74.7} & \textbf{99.6} \\
\bottomrule 
\end{tabular}
\\
\vspace{0.03cm}
\footnotesize{1. mIoU of objects; 2. mIoU of drivable area.}
\end{table}

\begin{figure*}
    \centering
    \includegraphics[width=0.99\linewidth]{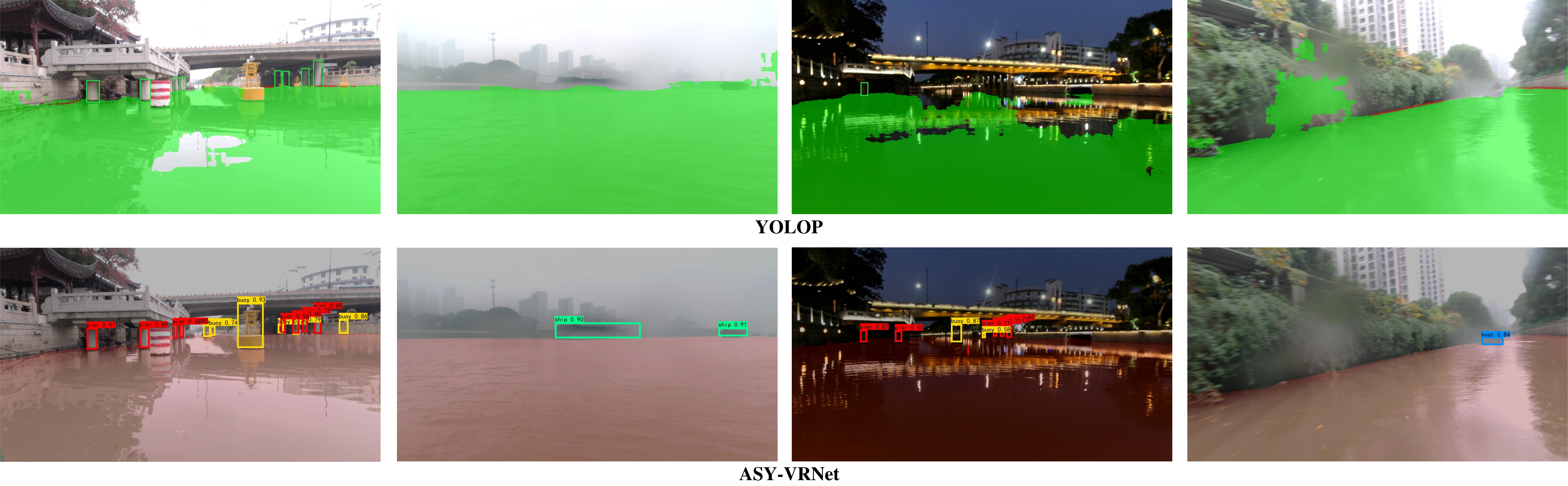}
    \vspace{-3mm}
    \caption{Visualization of panoptic driving perception results predicted by YOLOP and ASY-VRNet, including scenarios of dense objects, dense fog, low light and droplets on the lens. Besides, the drivable area predicted by YOLOP is green while ASY-VRNet's is red.}
    \label{fig:compare_yolop}
\end{figure*}

\subsection{Comparison on Object Detection}
As TABLE \ref{tab:detection_compare_map} shows, for object detection, we compare performances of single-task models, multi-task models, vision-based models, and radar-vision fusion models, which mainly include models with similar orders of magnitude of parameters. Generally speaking, our ASY-VRNet achieves state-of-the-art performances whatever mAP$_\text{50-95}$ and AR$_\text{50-95}$ with generally fewer parameters and FLOPs. Exactly, compared with another fusion-based PDP multi-task model Achelous (MV-GDP-X-PN), our ASY-VRNet exceeds about 1.3 mAP$_\text{50-95}$ while 0.7 AR$_\text{50-95}$. For another two vision-based multi-task models with more parameters and FLOPs, YOLOP and HybridNets, our ASY-VRNet achieves 3.7 and 4.9 mAP$_\text{50-95}$ higher than them. Furthermore, our ASY-VRNet gets the best performances when compared with single-task models. Notably, ASY-VRNet outperforms CRFNet 1 mAP$_\text{50-95}$ with about 19 million fewer parameters. From another perspective, we find that fusion-based models generally obtain better detection recall than vision-based models, which means a lower miss-detection rate based on fusion-based perception methods. Further, as TABLE \ref{tab:adverse_performance} shows, ASY-VRNet outperforms Achelous under several adverse situations.

\subsection{Comparison on Semantic Segmentation}
TABLE \ref{tab:detection_compare_miou} presents the semantic segmentation results of various models on the segmentation of objects and drivable areas. It is apparent that our ASY-VRNet achieves the best performance among all single-task models and multi-task models. Specifically, ASY-VRNet outperforms Achelous (MV-GDP-X-PN) by about 4.1 mIoU when segmenting objects. Although YOLOP and HybridNets have more parameters, their performances are still worse than ASY-VRNet, which proves the effectiveness of our model architecture. Besides, ASY-VRNet achieves 1.2 higher mIoU than Segformer-B0 with fewer FLOPs.

\subsection{Ablation Experiments}
To demonstrate the efficacy of our ASY-VRNet's modules, we conduct ablation experiments. As summarized in TABLE \ref{tab:ablation_experiments}, various modules including RIM, IRC, the fusion branch in the neck (neck fusion), CoC-FPN, decouple detection head, and multi-frame radar data are evaluated. Notably, for object detection, IRC has the most significant impact, leading to a decrease in mAP of approximately 1.2\%. Additionally, replacing CoC-FPN with a conventional convolutional FPN resulted in a 1.0\% drop in mAP. Furthermore, the neck fusion branch, decouple detection head, and multi-frame data each contributed to enhancing object detection performance to varying degrees. For semantic segmentation, RIM is found to be the most critical component. Notably, replacing CoC-FPN with a conventional FPN causes a 1.8\% decline in mIoU. CoC-FPN also demonstrates some improvement in semantic segmentation. Based on these findings, we observe that removing CoC-FPN hurt both object detection and semantic segmentation. This suggests that maintaining consistency between the feature structure of the FPN (decoder) and the backbone (encoder) is essential.

\begin{figure*}
    \centering
    \includegraphics[width=0.99\linewidth]{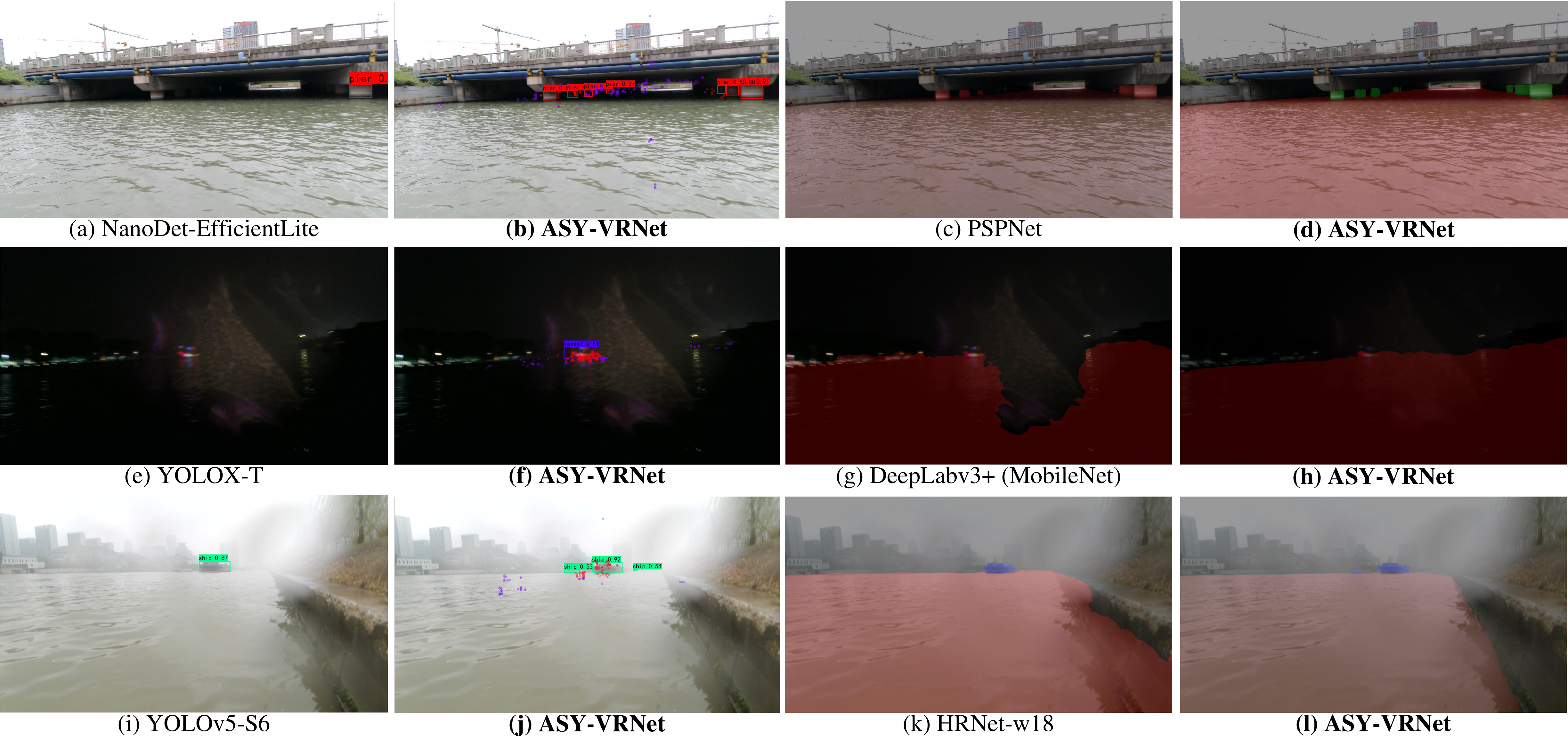}
    \vspace{-3mm}
    \caption{Visualization of single-task models and ASY-VRNet, presenting distant vessel at night and interfered camera, and multiple ships on a foggy day. The first column is results of pure vision models while the second column presents the detection of our ASY-VRNet with 4D radar point clouds. The third column presents the semantic segmentation of pure vision models while the fourth column shows the segmentation of ASY-VRNet.}
    \label{fig:all_compare}
\end{figure*}

\subsection{Comparison on Multi-Task Training}

As shown in TABLE \ref{tab:training_methods_compare}, we utilize four multi-task training techniques to train our ASY-VRNet. The techniques include task-joint methods containing uncertainty weighting, manual weighting, GradNorm \cite{chen2018gradnorm} and MGDA \cite{sener2018multi}. Notably, the uncertainty-based training approach delivers exceptional overall performance, with the lowest detection loss and optimized segmentation results. When training object detection independently, the mAP metric is unsatisfactory. It's noteworthy that manually tuning the weights of the four sub-tasks is a challenging task, as its optimization performance falls short of the uncertainty-based training method. Furthermore, GradNorm and MGDA achieve competitive performances, but are still worse than our uncertainty-based training approach. In summary, our multi-task training strategy for the PDP model can effectively improve the performance of each individual task.

\begin{table}
\setlength\tabcolsep{4.0pt}
\caption{Ablation Experiments of ASY-VRNet}
\vspace{-3mm}
\centering
\label{tab:ablation_experiments}
\begin{tabular}{l|lll}  
\toprule   
  \textbf{Methods} & \textbf{mAP$_\text{50-95}$} & \textbf{mIoU$_{\text{o}}$} & \textbf{mIoU$_{\text{d}}$} \\
\midrule   
  \textbf{ASY-VRNet} & \textbf{42.8} & \textbf{74.7} & \textbf{99.6} \\
\hline 
  -RIM & - & 72.9 ($\downarrow 1.8$) & -  \\
  -IRC & 41.6 ($\downarrow 1.2$) & - & \\
  -neck fusion & 42.4 ($\downarrow 0.4$) & - & \\
  -decouple detection head & 42.6 ($\downarrow 0.2$) & - & \\
  -SA\&CA on radar maps & 42.6 ($\downarrow 0.2$) & 74.4 ($\downarrow 0.3$) & 99.5 ($\downarrow 0.1$) \\
  multi-frame$\rightarrow$single-frame & 42.3 ($\downarrow 0.5$) & 74.1 ($\downarrow 0.6$) & - \\
  CoC-FPN $\rightarrow$ Conv-FPN & 41.8 ($\downarrow 1.0$) & 73.9 ($\downarrow 0.8$) & 99.1 ($\downarrow 0.5$) \\
\bottomrule  
\end{tabular}
\end{table}

\begin{table}
\setlength\tabcolsep{1.2pt}
\scriptsize
\caption{Results of Multi-Task Training Methods}
\vspace{-3mm}
\centering
\label{tab:training_methods_compare}
\begin{tabular}{c|cccc|cc|ccc}  
\toprule   
  \multirow{2}*{Methods} & \multicolumn{4}{|c|}{Weights} & \multirow{2}*{loss$_{\text{det}}$} & \multirow{2}*{loss$_{\text{seg}}$} & \multirow{2}*{\textbf{mAP$_\text{50-95}$}} & \multirow{2}*{\textbf{mIoU$_{\text{o}}$}} & \multirow{2}*{\textbf{mIoU$_{\text{d}}$}} \\
          & det$_{\text{bbox}}$ & det$_{\text{conf}}$ & det$_{\text{cls}}$ & seg$_{\text{cls}}$ &  & & \\
\midrule   
\textbf{Uncertainty}  & \multirow{2}*{\checkmark} & \multirow{2}*{\checkmark} & \multirow{2}*{\checkmark} & \multirow{2}*{\checkmark} & \multirow{2}*{\textbf{3.273}} & \multirow{2}*{\textbf{0.301}} & \multirow{2}*{\textbf{42.8}} & \multirow{2}*{\textbf{74.7}} & \multirow{2}*{\textbf{99.6}} \\
 \textbf{Weighting} &  &  &  &  &  & \\
\hline 
Manual & 0.6 & 0.2 & 0.2 & 0.0 & 3.917 & - & 42.6 & - \\
Manual & 0.5 & 0.2 & 0.2 & 0.1 & 3.578 & 0.355 & 42.3 & 74.4 & 99.6 \\
Manual & 0.25 & 0.25 & 0.25 & 0.25 & 4.231 & 0.322 & 42.0 & 73.9 & 99.3 \\
Manual & 0.1 & 0.2 & 0.2 & 0.5 & 5.241 & 0.352 & 41.6 & 74.6 & 99.6 \\
Manual & 0.0 & 0.0 & 0.0 & 1.0 & - & 0.303 & - & 74.6 & 99.6 \\
\midrule
GradNorm \cite{chen2018gradnorm} & - & - & - & - & 3.327 & 0.349 & 42.5 & 74.4 & 99.6 \\
MGDA \cite{sener2018multi} & - & - & - & - & 3.319 & 0.356 & 42.3 & 74.3 & 99.6 \\
\bottomrule  
\end{tabular}
\end{table}

\subsection{Experiments on Fusion Methods}
We compare our proposed Asymmetric Fair Fusion (AFF) modules with several well-known fusion methods, including Multi-Head Cross Attention (MHCA), TokenFusion, the fast fusion module in Achelous \cite{guan2023achelous}, and Dual Graph Fusion (DGF) \cite{guan2024mask}. As shown in TABLE \ref{tab:fusion_experiments}, our AFF modules achieve state-of-the-art performance across three perception tasks, surpassing other fusion methods. The performance of MHCA, which relies on the global receptive field, is not satisfactory. Similarly, the modal-agnostic fusion method TokenFusion shows a significant performance gap compared to our AFF modules, underscoring the importance of dedicated fusion methods tailored for various tasks.

\begin{table}
\setlength\tabcolsep{12.0pt}
\caption{Comparison of Fusion Methods for Vision and Radar}
\vspace{-3mm}
\centering
\label{tab:fusion_experiments}
\begin{tabular}{l|lll}  
\toprule   
  \textbf{Methods} & \textbf{mAP$_\text{50-95}$} & \textbf{mIoU$_{\text{o}}$} & \textbf{mIoU$_{\text{d}}$} \\
\midrule   
  \textbf{AFF (ours)} & \textbf{42.8} & \textbf{74.7} & \textbf{99.6} \\
\hline 
  MHCA \cite{wu2023referring} & 40.4 & 68.2 & 97.5 \\
  TokenFusion \cite{wang2022multimodal} & 41.2 & 68.0 & 97.8  \\
  Achelous \cite{guan2023achelous} & 42.1 & 72.0 & 99.5 \\
  DGF \cite{guan2024mask} & 42.3 & 72.8 & 99.4 \\
\bottomrule  
\end{tabular}
\end{table}

\begin{table}
\setlength\tabcolsep{2.3pt}
\caption{Comparison of Models under Adverse Situations}
\vspace{-3mm}
\centering
\label{tab:adverse_performance}
\begin{tabular}{l|cc|cc|cc}  
\toprule   
  \textbf{Models} & \textbf{mAP$^{\text{da}}$} & \textbf{mIoU$^{\text{da}}_{\text{d}}$} & \textbf{mAP$^{\text{di}}$} & \textbf{mIoU$^{\text{di}}_{\text{d}}$} & \textbf{mAP$^{\text{sm}}$} & \textbf{mIoU$^{\text{sm}}_{\text{o}}$} \\
\midrule   
  \textbf{ASY-VRNet} & \textbf{38.8} & \textbf{93.7} & \textbf{39.5} & \textbf{95.6} & \textbf{36.7} & \textbf{68.8} \\
\hline 
  Achelous & 37.2 & 90.9 & 38.8 & 95.2 & 33.0 & 63.7 \\
\bottomrule  
\end{tabular}\\
\footnotesize{\textbf{da}: dark, \textbf{di}: dim, \textbf{sm}: small, \textbf{d}: drivable-area, \textbf{t}: target}
\end{table}

\subsection{Visualization and Analysis}
We visualize the prediction results of our ASY-VRNet and other models as Fig. \ref{fig:compare_yolop} and Fig. \ref{fig:all_compare} present. We first select four representative samples under various scenarios, including dense and small objects, dense fog, low light and droplets on the lens. For the scenario containing dense small objects, we can find that our ASY-VRNet can nicely detect all objects with high confidence scores while YOLOP misses considerable objects. For the objects behind the dense fog, YOLOP, unfortunately, misses both two ships while our ASY-VRNet successfully detects two moving ships. Moreover, the result of drivable-area segmentation by ASY-VRNet is better than YOLOP. For the scenario of low light, the drivable area predicted by YOLOP contains a lot of false-negative zones while our ASY-VRNet can better recognize them. For the fourth sample, we find that YOLOP can not perceive the object when some water droplets on the camera occluded the object, and it predicts many false positive drivable areas. In contrast, our ASY-VRNet can still capture the driving boat and smoothly segment the driving area. From another perspective, when compared with single-task models in Fig. \ref{fig:all_compare}, we can also observe the outstanding performances of ASY-VRNet in different scenarios.

\section{Conclusions}
\label{sec:conclusions}
In this paper, we propose ASY-VRNet, a Panoptic Driving Perception (PDP) model for waterways that concurrently performs two distinct tasks. Our Asymmetric Fair Fusion (AFF) module efficiently integrates complementary features from each modality, enhancing the performance of both tasks simultaneously. Furthermore, we adopt a homoscedastic-uncertainty-based multi-task training method tailored for panoptic perception tasks, demonstrating its efficacy. ASY-VRNet treats both images and radar point clouds as irregular point sets, achieving competitive performance compared to other state-of-the-art single-task models. Moreover, it surpasses existing vision-based and fusion-based PDP models in overall performance.

\section*{Acknowledgment}

This work is partially supported by the XJTLU AI University Research Centre and Jiangsu Province Engineering Research Centre of Data Science and Cognitive Computation at XJTLU. Also, it is partially funded by the Suzhou Municipal Key Laboratory for Intelligent Virtual Engineering (SZS2022004) as well as funding: XJTLU-REF-21-01-002, XJTLU- RDF-22-01-062, and XJTLU Key Program Special Fund (KSF-A-17). This work received financial support from Jiangsu Industrial Technology Research Institute (JITRI) and Wuxi National Hi-Tech District (WND).

\footnotesize
\bibliographystyle{ieeetr}
\bibliography{refs}

\end{document}